# Multimodal Semantic Simulations of Linguistically Underspecified Motion Events

Nikhil Krishnaswamy and James Pustejovsky

Department of Computer Science, Brandeis University
415 South Street, Waltham, MA 02453, USA
{nkrishna,jamesp}@brandeis.edu
http://www.voxicon.net/

**Abstract.** *In this paper, we describe a system for generating three-dimensional visual simulations of natural language motion expressions. We use a rich formal model of events and their participants to generate simulations that satisfy the minimal constraints entailed by the associated utterance, relying on semantic knowledge of physical objects and motion events. This paper outlines technical considerations and discusses implementing the aforementioned semantic models into such a system.*

## 1 Introduction

Existing work in visualization from natural language has largely focused on object placement in static scenes [3, 5, 31]. By focusing on motion verbs, in this paper we outline an approach to integrating dynamic semantics into visualization, resulting in simulations of the associated actions. In philosophy, "mental simulation" theory attempts to model everyday human psychological competence [19], providing a *process* driven theory of mind [13]. In cognitive linguistics, "simulation" has come to mean a mental instantiation of a linguistic utterance, playing a functional role in language understanding [7, 1], based on the notion of an agent's *embodiment* [23]. Finally, both QSR and gaming-style AI approaches have been used to develop scenario-based simulators, such as for trainers driven by interactive narratives [8, 6].

In a dynamic semantics approach, verbs are treated as programs or processes [17] and so although the computational linguistics and cognitive linguistics communities do not often reference each other, in our opinion there is fertile ground for cross-pollination, starting with the approach of Pustejovsky and Moszkowicz [29], and to implement language-based reasoning in a QSR framework.

We previously presented a method for visualizing natural language expressions in a 3D environment built on the Unity game engine [27]. This was followed by the development of VoxML [28], a modeling language which encodes object and event semantic information into *voxemes* or "visual object concepts," and a *voxicon* (the "lexicon" of voxemes). This approach enables procedural simulation generation from semantic knowledge of an event and its participants. Our system, given an utterance and scene containing all referenced nominals as 3D objects, enacts the verbal program over them. The remainder of this paper describes in brief the workflow of this system, the Voxicon simulator.

## 2 Architecture

The software uses the Unity game engine [14] for graphics and I/O processing. Input is a simple natural language sentence, which is part-of-speech tagged and dependency-parsed. These NLP tasks are currently handled by external processors (such as the ClearNLP parser [4]) networked to the simulator. 3D assets and VoxML-modeled nominal objects and events (created with other Unity-based tools) are loaded externally, either locally or from a web server.

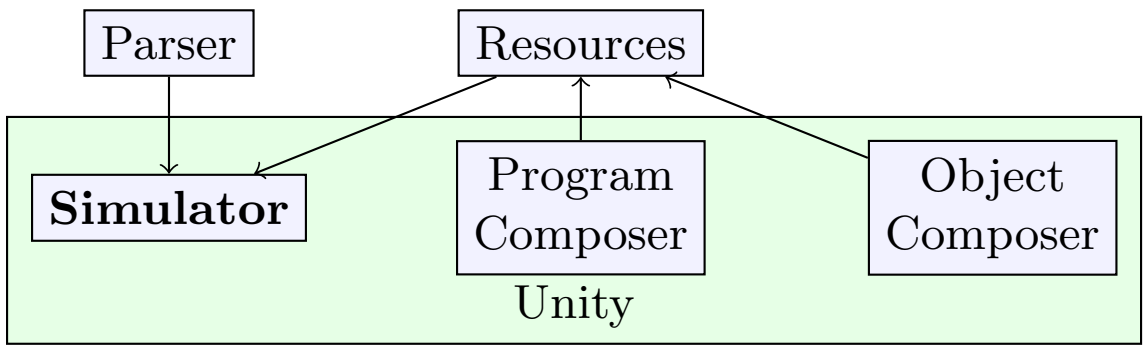

**Fig. 1.** Voxicon architecture schematic

## 3 Linguistic and Semantic Analysis

From tagged and parsed input text, all noun phrases are cross-referenced to objects in the scene. A reference to *ball* causes the simulator to attempt to locate a "ball" voxeme instance in the scene while *table* prompts an attempt to locate a "table" voxeme instance. Attributive adjectives impose a sortal scale on their heads, so *small table* and *big table* link to two separate tables if they exist in the scene, and the VoxML-encoded semantics of "small" and "big" discriminates the tables based on their relative size [28].

We use a basic set of primitive programs to represent verbs, from which we build more complex programs. There have been many previous attempts to group verbs into distinct clusters, based on syntactic behavior [16], or to associate verbs with specific spatial semantic primitives [22]. Frameworks from the computer vision and AI communities include work on representing traffic occurrences [11] and in human-robot communication [32]. VoxML follows a similar model-theoretic approach, and while VoxML and voxeme primitive programs have an underlying semantics of a hybrid dynamic logic [29], their content is largely based on implementation considerations, and focus on decomposing the verbal and relational semantics while leaving object category labels intact, similar to approaches that seek to overcome descriptive constraints limited by robotic perception [21].

All 3D motion can be decomposed into translations and rotations, making those obvious verbal primitives in our system. Others include commonly-repeating subevents of other motions, such as "grasp," a fine-grained motion that is difficult to decompose but appears as a subevent of nearly any human-object interaction. We can then assemble complex events out of primitive motions, as in "put" in Fig. 2 below, and then macro-complex events, such as "stack" as a sequence of "put" events.

### 3.1 Habitats and Affordances

We assume that every voxeme exists within an intrinsic "habitat" [25, 20], an encoding of the environment in which the object must exist simply to avoid violating any physical constraints, such as gravity, and conditions under which an object typically exists in the world. For instance, a pencil must be laying flat and not resting on its tip. A table can be situated in almost any orientation, but has an intrinsic "top" (its surface). This is typically oriented along the world's upward vector, so we denote this in VoxML with TOP = $top(+Y)$.

A voxeme's semantic structure also provides "Gibsonian" and "telic" *affordances* [12, 24, 26], or attached behaviors, which the object either facilitates by its geometry, or purposes for which it is intended to be used. For example, a Gibsonian affordance for a cup may be "grasp," while a telic affordance may be "drink from."

### 3.2 Predicate-Argument Interaction

VoxML treats objects (NPs) and events (predicates) in terms of a dynamic event semantics, Dynamic Interval Temporal Logic (DITL) [29]. The verbal semantics follow a type system that encodes how a given formula $\phi$ and proposition $\pi$ are executed/tested during the execution of a verbal program; programs can be a `state`, `process`, `transition`, `assignment`, or `test`, all of which are translated into DITL and operationalized differently.

Adopting a dynamic interpretation of events allows us to map linguistic expressions directly into simulations through an operational semantics. Further, predicates are interpreted relative to their arguments' semantic encoding.

In Fig. 2, "put" is given arguments `agent`, `obj1`, and `on(obj2)`. The typing of both "put" and "on" encode the calculation of such parameters as object trajectory and destination location (e.g. the position denoted by `on(block)` vs `on(plate)` or `on(wall)`). As seen in BODY, object $A_2$ is moved to location $A_3$, calculated by operationalizing `on` over `obj2`. The results, depending on the arguments, may be configurations such as those shown in Figs. 3-5.

$$
\left[\begin{array}{l}
\textbf{put} \\
\text{LEX} = \left[\begin{array}{lcl} \text{PRED} & = & \textbf{put} \\ \text{TYPE} & = & \textbf{transition_event} \end{array}\right] \\
\text{TYPE} = \left[\begin{array}{ll}
\text{HEAD} = & \textbf{transition} \\
\text{ARGS} = & \left[\begin{array}{lcl} \text{A}_1 & = & \textbf{agent} \\ \text{A}_2 & = & \textbf{obj1} \\ \text{A}_3 & = & \textbf{on(obj2)} \end{array}\right] \\
\text{BODY} = & \left[\begin{array}{lcl} \text{E}_1 & = & grasp(A_1, A_2) \\ \text{E}_2 & = & [while(hold(A_1, A_2), move(A_2))] \\ \text{E}_3 & = & [at(A_2, A_3) \rightarrow ungrasp(A_1, A_2)] \end{array}\right]
\end{array}\right]
\end{array}\right]
$$

**Fig. 2.** VoxML structure for "put"

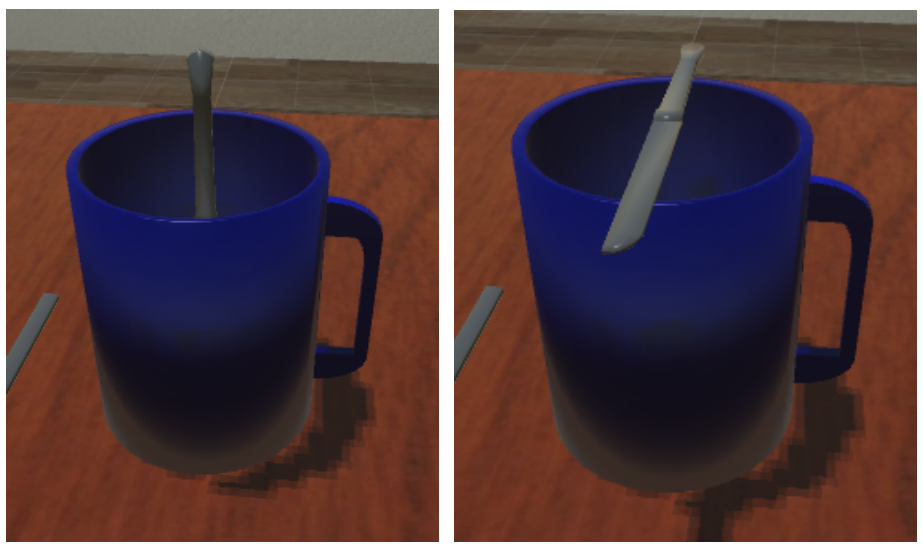

**Fig. 3.** "In the cup" vs. "on the cup"

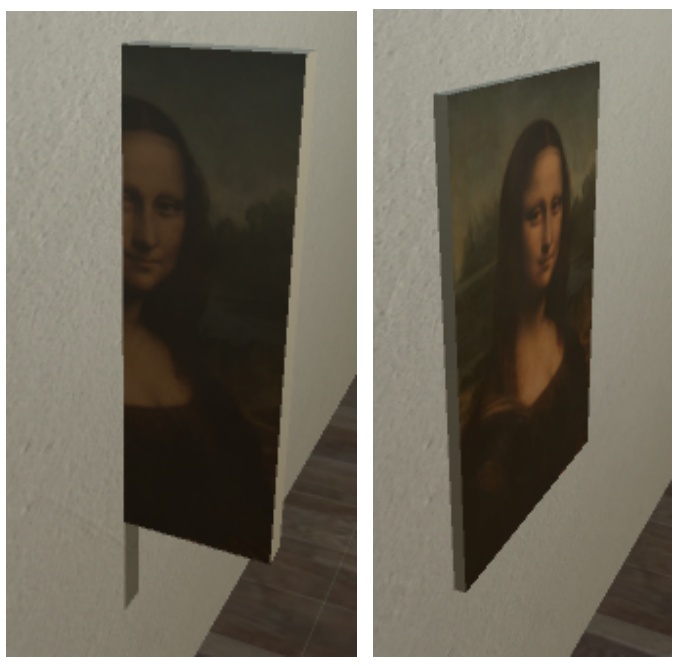

**Fig. 4.** "In the wall" vs. "on the wall"

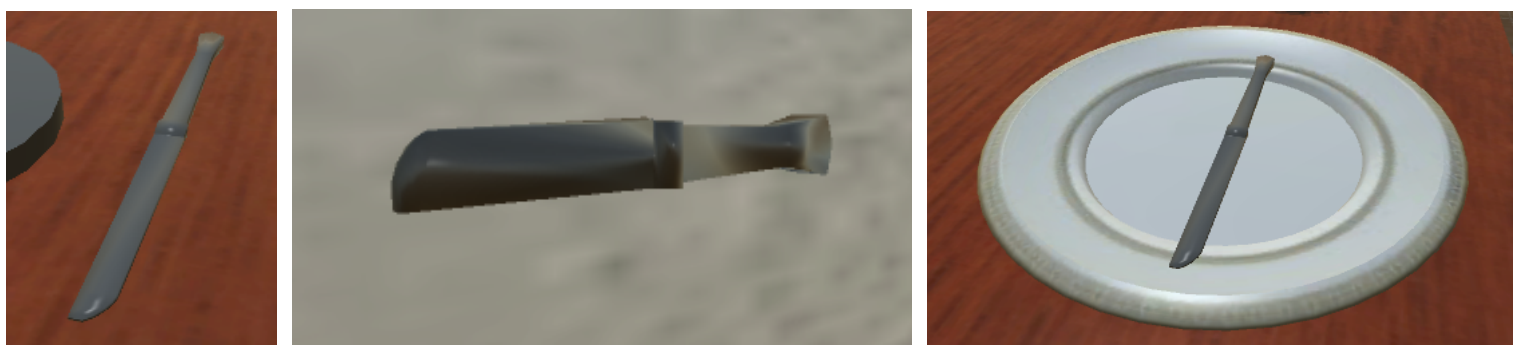

**Fig. 5.** "On the table" vs. "on the wall" vs. "on the plate"

Argument-sensitive distinctions in the operationalization of predicates themselves exploit the HEAD of the argument's type structure—a selected surface wholly or partially coterminous with the entire object.

For example, `on(wall)` selects for a vertical face of the object while in all other examples, `on` selects for the object's top. The location of "top" is computed based on the object's VoxML type structure. The top of a plate, a slightly concave object when situated in its default habitat, is located slightly lower on the Y-axis than the plate's highest point. `in(cup)` selects the interior surface of the cup, an object with a telic affordance of containment, while `in(wall)` explicitly interpenetrates the wall, as "wall" lacks the containment telic role. The system attempts to maximally satisfy the constraints placed upon it by the NL expression. For instance, placing the knife object `on(cup)` lays the knife across the rim of the upright cup. In the same (horizontal) orientation, the knife cannot be placed at the location computed by `in(cup)`, so the system must transform the knife so that the RCC representation of `in(cup)`, $PO(knife, cup)$, can be satisfied without violating any of the physical or structural constraints of the cup or knife objects.

Given any object pair, the system can pinpoint the set of RCC relations applying to them in the current state. RCC8 relations in 3D may leave the dimensionality of the relation unspecified, so VoxML picks up here.[1] For example, in VoxML a "cup" is a concave object with reflectional symmetry across the XY and YZ planes, and with rotational symmetry around the Y axis. The symmetry information entails that the concavity must open along the Y axis while the habitat information entails that it opens toward the top of the cup in default orientation. The cup's affordance structure includes containment, so any object to be placed "in" the cup must be tested to see if it fits inside when aligned to the cup's Y axis along its longest dimension.

## 4 Discussion and Future Work

We have presented here a method for incorporating motion and dynamic spatial semantics into a visualization framework. Generating a visualization at runtime necessitates some constraints left unspecified in a minimal model be made explicit in the simulation. These include direction in a bare manner-of-motion verb (e.g. "the ball rolled"). We are developing a Monte-Carlo method to determine prototypical values for these constraints, as well as evaluation methods for the results of these experiments. In instances where a program's DITL formula says nothing about the nature of a given parameter (e.g. states that $b_{n+1}$ is farther from $b_0$ than $b_n$ is but does not state in which direction), this Monte-Carlo method should allow us to either compute a "prototypical value" for that parameter, or to state that none seems evident. We are exploring three evaluation methods to determine which variable values occur in the simulation(s) judged the best for a given description label:

1. Given a visualization and a set of potential descriptions, use Amazon Mechanical Turk to gather the best description.
2. Given a description and a set of possible simulations, gather the best simulation.
3. Using operationalizations of corpus-derived verb satisfaction conditions, automatically compute vector distance from a simulation's final state to the satisfaction condition.

[1] The currently-implemented reasoning approach relies on RCC8 [30, 9, 10], but can easily be extended to other QSR approaches, including the situation calculus [2], and the Intersection Calculus [18, 15].

Finally, we are also developing methods for automatically composing complex behaviors from primitives, based on Dynamic Interval Temporal Logic [29] as well as building a corpus of linked simulations and event-annotated video in order to train algorithms to discriminate events based on their participants' motions.

## Acknowledgements

This work is supported by a contract with the US Defense Advanced Research Projects Agency (DARPA), Contract W911NF-15-C-0238. Approved for Public Release, Distribution Unlimited. The views expressed are those of the authors and do not reflect the official policy or position of the Department of Defense or the U.S. Government. All errors and mistakes are, of course, the responsibilities of the authors.